\documentclass[11pt]{article}

\usepackage[preprint]{acl}
\usepackage{times}
\usepackage{latexsym}
\usepackage[T1]{fontenc}
\usepackage[utf8]{inputenc}
\usepackage{microtype}
\usepackage{inconsolata}
\usepackage{graphicx}
\usepackage{booktabs}
\usepackage{amsmath}
\usepackage{amssymb}
\usepackage{bm}
\usepackage{enumitem}
\usepackage{multirow}
\usepackage{hyperref}
\usepackage[table]{xcolor}
\usepackage{tabularx}
\usepackage{array}
\usepackage{algorithm}
\usepackage{algpseudocode}

\usepackage[most]{tcolorbox}
\usepackage{listings}

\definecolor{caseaccent}{HTML}{4A7C9B}    
\definecolor{casebg}{HTML}{F6F8FA}        
\definecolor{casetitle}{HTML}{2C3E50}     

\definecolor{prompttitlebg}{HTML}{EDEDED}
\definecolor{promptframe}{HTML}{555555}
\definecolor{placeholdercolor}{HTML}{333333}
\definecolor{jsonbg}{HTML}{F7F7F7}

\newcommand{\placeholder}[1]{%
  \texttt{\{#1\}}%
}

\newtcolorbox{simplepromptbox}[1]{
  enhanced,
  colback=white,
  colframe=promptframe,
  colbacktitle=prompttitlebg,
  coltitle=black,
  title=\textbf{#1},
  fonttitle=\small\bfseries,
  boxrule=0.5pt,
  arc=0pt,
  left=1.5mm,
  right=1.5mm,
  top=1mm,
  bottom=1mm,
  boxsep=0pt,
  before skip=4pt,
  after skip=4pt,
  before upper={%
    \setlength{\parskip}{2pt}%
    \setlength{\parindent}{0pt}%
  }
}

\lstdefinestyle{promptjsonstyle}{
  basicstyle=\ttfamily\scriptsize,
  breaklines=true,
  columns=fullflexible,
  keepspaces=true,
  showstringspaces=false,
  frame=none,
  backgroundcolor=\color{jsonbg},
  xleftmargin=0.5em,
  xrightmargin=0.5em
}

\newtcolorbox{evolutioncase}[2][]{%
  enhanced,
  breakable,
  colback=white,
  colframe=promptframe,
  colbacktitle=prompttitlebg,
  coltitle=black,
  fonttitle=\small\bfseries,
  title={#2},
  boxrule=0.5pt,
  arc=0pt,
  left=1.5mm, right=1.5mm, top=1mm, bottom=1mm,
  boxsep=0pt,
  before upper={%
    \setlength{\parskip}{2pt}%
    \setlength{\parindent}{0pt}%
  },
  #1
}

\lstnewenvironment{promptjson}
{\lstset{style=promptjsonstyle}}
{}

\title{EDGE: Experience-Distillation for Guided Exploration in Agentic Reinforcement Learning}

\author{
\textbf{
Can Xie\textsuperscript{1,2},
Yuyi Zhou\textsuperscript{2,3},
Wen Yang\textsuperscript{1,2},
Ziyi Zhang\textsuperscript{2,3},
}\\
\textbf{
Siyao Song\textsuperscript{1,2},
Yingzhuo Deng\textsuperscript{1,2},
Shuo Ren\textsuperscript{2}\thanks{Corresponding authors},
Jiajun Zhang\textsuperscript{1,2,4}\footnotemark[1]
}, \\[5pt]
\textsuperscript{1}~School of Artificial Intelligence, University of Chinese Academy of Sciences\\
\textsuperscript{2}~Institute of Automation, Chinese Academy of Sciences \\
\textsuperscript{3}~School of Advanced Interdisciplinary Sciences, University of Chinese Academy of Sciences \\
\textsuperscript{4}~Wuhan AI Research \\
\texttt{\{xiecan2024,shuo.ren\}@ia.ac.cn}, \texttt{jjzhang@nlpr.ia.ac.cn}
}

\begin{document}
\maketitle

\begin{abstract}
Reinforcement learning with outcome-based objectives such as GRPO
enables LLM-based agents to solve complex, long-horizon tasks,
yet the reusable exploration patterns embedded in interaction
trajectories are largely discarded after a single policy update.
Existing experience-augmented approaches retrieve historical
guidance at inference time, but they apply experiences without
accounting for the policy's evolving capability and create
persistent dependencies on external retrieval.
We propose \textbf{EDGE}
(\textbf{E}xperience-\textbf{D}istillation for
\textbf{G}uided \textbf{E}xploration), a framework that
treats retrieved experiences as temporary training-time scaffolds
and progressively internalizes their benefits into the parametric
policy.
Concretely, EDGE partitions each rollout group into experience-conditioned and experience-free trajectories to estimate and admit only positive marginal gains without extra sampling, then distills the induced behavior into the base policy via a reverse-KL objective on its own empirical support. A co-evolutionary experience bank further synthesizes guidance from emerging failure modes and prunes obsolete entries as the policy evolves.
Across embodied, web, and search-based QA tasks, EDGE improves over strong
RL baselines by up to 12.5 points and remains effective without
inference-time scaffolds or a proprietary reflector. The code is available at \url{https://github.com/xvolcano02/EDGE}.
\end{abstract}

\newcommand{\methodname}{\textsc{EDGE}}

\begin{figure*}[t] 
\centering
    \includegraphics[width=0.95\textwidth]{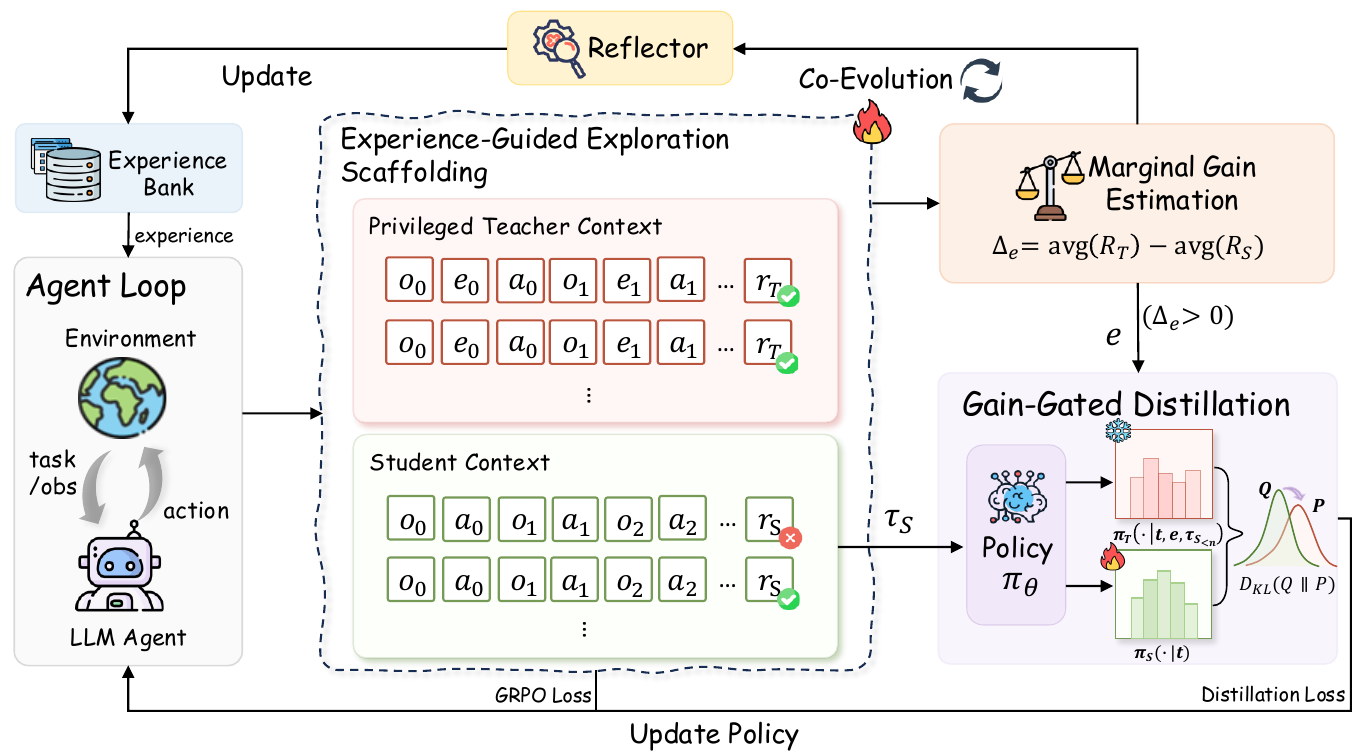}
\caption{\textbf{Overview of the EDGE framework.} (a) Experience-guided exploration scaffolding uses contrastive rollouts to estimate the marginal gain $\Delta_e$ of a retrieved experience. (b) Gain-gated distillation internalizes beneficial scaffold behavior into the base policy only when $\Delta_e > 0$. (c) The experience bank co-evolves with the policy via failure-driven expansion and utility-based pruning, yielding a retrieval-free policy at deployment.}
\label{fig:overview}
\end{figure*}

\section{Introduction}\label{sec:intro}

Reinforcement learning (RL) with outcome-based objectives such as
GRPO~\citep{shao2024deepseekmath} has become a standard post-training
paradigm for LLM-based agents that reason, plan, and act through multi-turn
interactions~\citep{yao2023react,shinn2024reflexion,DBLP:conf/nips/Yao0YN22,DBLP:conf/iclr/ShridharYCBTH21,feng2025gigpo,wang2025ragen,jin2025searchr1}.
Yet current agentic RL uses its own experience poorly.
A single rollout may contain reusable patterns---effective task
decompositions, recovery strategies, dead-end avoidance, but once a
trajectory contributes a scalar advantage to one policy update, these
patterns are largely discarded.
The agent must therefore rediscover them from scratch, a particularly costly
failure mode in sparse-reward, long-horizon environments where agentic RL is
most needed.

A growing body of work addresses this inefficiency by augmenting agents with
external experience at inference time, through episodic
reflections~\citep{shinn2024reflexion,zhao2024expel}, persistent
memory~\citep{chhikara2025mem0,fang2026mempexploringagentprocedural}, or skill
libraries~\citep{xia2026skillrl,liu2026exploratorymemoryaugmentedllmagent}.
While effective, these approaches share two structural limitations that
become apparent as training progresses.
Experience utility is inherently policy-dependent: guidance that accelerates
an undertrained policy can become redundant---or actively harmful---once the
corresponding behavior has been internalized, yet most methods apply
experience unconditionally or filter it by static heuristics.
Moreover, if the agent must retrieve experience at deployment, part of its
competence resides in the context window rather than in the model parameters,
incurring persistent token overhead and sensitivity to retrieval noise. \footnote{These are not hypothetical concerns: in our experiments, naive memory
augmentation (e.g., EvolveR, GRPO+Mem0) degrades performance well below
vanilla GRPO, confirming that unfiltered experience injection is
unreliable.}
These observations suggest that experience reuse in agentic RL should be treated as a dynamic lifecycle rather than a static retrieval mechanism.
External experience could first guide exploration, as a temporary training-time scaffold, then be exploited by consolidating its useful behavioral effect into the experience-free policy, and finally be retired once its utility vanishes.
Under this view, experience reuse becomes a scaffold-to-parameter learning problem rather than a retrieve-and-prompt augmentation problem.


In this paper, we instantiate this idea in EDGE (Experience-Distillation for Guided Exploration), a framework that turns retrieved experience from persistent inference-time memory into dynamically validated training-time scaffolding. EDGE uses online marginal-gain estimation to decide when an experience should guide exploration, be distilled into the policy, or be retired as the policy evolves, via three core mechanisms. 
First, \textbf{experience-guided exploration scaffolding} partitions each GRPO rollout group into experience-conditioned and experience-free trajectories, estimating the marginal gain of a retrieved experience without additional environment sampling and retaining only positive-gain instances. 
Then, \textbf{gain-gated privileged distillation} serves as the exploitation step, transferring the behavior induced by the privileged context into the standard policy via reverse-KL divergence computed on the student's own empirical support, activating only when the estimated gain is positive to avoid negative transfer. 
Finally, \textbf{experience bank evolution and management} enables the experience bank to co-evolve with the policy throughout training: new experiences are synthesized from failure-mode analysis, while obsolete ones are pruned based on tracked utility, keeping the scaffold aligned with the agent's evolving capability.

On ALFWorld~\citep{DBLP:conf/iclr/ShridharYCBTH21} and
WebShop~\citep{DBLP:conf/nips/Yao0YN22}, \methodname{} consistently
outperforms skill-free RL and prior experience-augmented methods, with the
largest gains on exploration-intensive subtasks (e.g., Heat, Cool, Pick2).
When external experiences are withheld at inference time, \methodname{}
retains 96.0\% of its scaffolded performance---compared with 82.9\% for
SkillRL and 92.3\% for EMPO$^2$---confirming that useful exploration priors
have been absorbed into the parametric policy. On seven search-based QA
benchmarks with Qwen3-4B, \methodname{} further improves over Search-R1 by
5.9 points on average; using the policy itself as the reflector preserves
97.3\% of the performance obtained with GPT-4o.

Our contributions are as follows:
\begin{itemize}[leftmargin=1.2em,itemsep=2pt]
  \item We identify two failure modes of experience-augmented agentic
    RL---policy-dependent experience utility and persistent inference-time
    retrieval dependence---and reframe experience reuse as a dynamic
    scaffold-to-parameter transition.
  \item We introduce experience-guided exploration scaffolding, which
    partitions rollout groups to estimate the marginal value of each
    retrieved experience under the current policy without requiring
    additional environment sampling.
  \item We propose gain-gated privileged distillation, which internalizes
    scaffold-induced behavior via reverse-KL on the student's own empirical
    support, gated by the estimated marginal gain to prevent negative
    transfer, together with a co-evolutionary experience bank that
    expands and prunes in response to the policy's evolving needs.
  \item Experiments across embodied, web, and search-based QA tasks show
    that \methodname{} improves task performance and training efficiency,
    transfers to a newer Qwen3 backbone, and remains effective with a
    self-reflector and without experience retrieval at deployment.
\end{itemize}

\section{Preliminaries}\label{sec:prelim}

We formalize the multi-turn decision-making process of an LLM-based agent
as a Partially Observable Markov Decision Process
(POMDP)~\citep{kaelbling1998planning}
$\langle \mathcal{S}, \mathcal{A}, \mathcal{O}, \mathcal{T},
\mathcal{R} \rangle$.
The observation space $\mathcal{O}$ consists of natural-language strings
emitted by the environment, and the action space $\mathcal{A}$ comprises
token sequences generated autoregressively by the policy $\pi_\theta$.
Given a task instruction $x$, the agent interacts with the environment over
a sequence of turns: at step $t$ it receives observation
$o_t \in \mathcal{O}$ and produces action $a_t \in \mathcal{A}$ according to
\begin{equation}\label{eq:policy}
  a_t \sim \pi_\theta(\cdot \mid x, h_t, o_t),
\end{equation}
where $h_t = (o_1, a_1, \ldots, o_{t-1}, a_{t-1})$ is the interaction
history.
An episode terminates upon task completion or at a maximum step limit,
yielding a sparse binary reward $R(\tau) \in \{0, 1\}$ and a full trajectory
\begin{equation}\label{eq:traj}
  \tau = (x,\; o_1, a_1,\; \ldots,\; o_T, a_T,\; R(\tau)).
\end{equation}

We build on Group Relative Policy Optimization
(GRPO)~\citep{shao2024deepseekmath}, which samples a group of $G$ parallel
trajectories $\{\tau_i\}_{i=1}^{G}$ per task and computes group-normalized
advantages:
\begin{equation}\label{eq:grpo_adv}
  \hat{A}_i = \frac{R(\tau_i) - \mathrm{mean}(\{R(\tau_j)\}_{j=1}^{G})}
  {\mathrm{std}(\{R(\tau_j)\}_{j=1}^{G})}.
\end{equation}
The policy is updated by maximizing a clipped surrogate objective with KL
regularization:
\begin{align}\label{eq:grpo}
  \mathcal{L}_{\text{RL}}(\theta) = -\mathbb{E}\!\Bigg[
    \frac{1}{G}\sum_{i=1}^{G}\frac{1}{|\tau_i|}
    \sum_{t=1}^{|\tau_i|}
    \Big(
    \min\!\big(
      \rho_{i,t}\,\hat{A}_i,\; \nonumber \\
      \mathrm{clip}(\rho_{i,t}, 1\!-\!\epsilon, 1\!+\!\epsilon)\,
      \hat{A}_i
    \big)
    - \beta\, D_{\mathrm{KL}}\big(\pi_\theta \| \pi_{\text{ref}}\big)
    \Big)
  \Bigg],
\end{align}
where $\rho_{i,t} =
\pi_\theta(a_t \mid x, h_t, o_t)\,/\,
\pi_{\theta_{\text{old}}}(a_t \mid x, h_t, o_t)$ is the importance
sampling ratio, $\pi_{\text{ref}}$ is the reference policy, and $\beta$
controls KL penalty strength.
By contrasting outcomes within each group, GRPO steers $\theta$ toward
successful action sequences without a learned value function.
However, in sparse-reward, partially observable environments, unguided
exploration often yields groups in which few or no trajectories succeed,
rendering the advantage estimate uninformative---a limitation we address
in the following section.

\section{Method: EDGE}\label{sec:method}

We present \textbf{EDGE} (\textbf{E}xperience-\textbf{D}istillation for
\textbf{G}uided \textbf{E}xploration), a framework that iteratively
strengthens the multi-turn reasoning capability of LLM-based agents
(Figure~\ref{fig:overview}).
Central to our approach is treating retrieved external experience not as a
static inference-time prompt---which inflates the context window and creates
persistent retrieval dependence---but as a \emph{training-time scaffold} that
guides exploration and is discarded at deployment.
The agent first explores with privileged access to experience, then distills
only the empirically beneficial behavior into its own parameters, so the
deployed policy requires no external scaffold.
Section~\ref{sec:scaffold} introduces the experience-guided exploration
scaffolding, Section~\ref{sec:distill} details the gain-gated privileged
distillation mechanism, and Section~\ref{sec:bank} describes the experience
bank evolution and management strategy.

\subsection{Experience-Guided Exploration Scaffolding}\label{sec:scaffold}

To provide exploratory guidance in the sparse-reward, partially observable
environments typical of agentic tasks, EDGE introduces a controlled
information asymmetry within the standard GRPO rollout group.
Given a task instruction $x$ and an experience bank
$\mathcal{E}$, we retrieve the top-$m$ most relevant
experiences by embedding similarity, using the task
instruction and initial environmental observations as
the query, and select the highest-scoring entry
$e \in \mathcal{E}$.
Recall that GRPO samples a group of $G$ parallel trajectories per task to
estimate relative advantages (Eq.~\eqref{eq:grpo_adv}).
We partition this group into two equal subsets without adding extra rollouts:
\begin{itemize}[leftmargin=1.2em,itemsep=2pt]
  \item \textbf{Teacher rollouts} ($\mathcal{T}^{\mathsf{T}}$, $G/2$
    trajectories): conditioned on the \emph{privileged context}
    $c^{\mathsf{T}} = x \oplus e$, where $\oplus$ denotes concatenation
    under a unified chat template.
  \item \textbf{Student rollouts} ($\mathcal{T}^{\mathsf{S}}$, $G/2$
    trajectories): conditioned on the \emph{standard context}
    $c^{\mathsf{S}} = x$ alone.
\end{itemize}
Because both subsets share the same policy $\pi_\theta$ and differ only in
whether the retrieved experience is visible, any performance gap can be
attributed to the informational advantage provided by $e$.
We quantify this gap via the \emph{instantaneous marginal gain}:
\begin{equation}\label{eq:gain}
  \Delta_e \;=\; \frac{1}{|\mathcal{T}^{\mathsf{T}}|}
  \sum_{\tau \in \mathcal{T}^{\mathsf{T}}} R(\tau)
  \;-\; \frac{1}{|\mathcal{T}^{\mathsf{S}}|}
  \sum_{\tau \in \mathcal{T}^{\mathsf{S}}} R(\tau),
\end{equation}
where $R(\tau)$ is the binary outcome reward defined in \S\ref{sec:prelim}.
A positive $\Delta_e$ signals that the experience provides useful guidance
beyond the agent's current capability; a non-positive value indicates it
is redundant or harmful.
This estimate gates the distillation objective (\S\ref{sec:distill}) and
drives experience bank updates (\S\ref{sec:bank}).
 
Beyond gating distillation, $\Delta_e$ controls which rollouts enter the RL
loss itself.
When $\Delta_e > 0$, advantages (Eq.~\eqref{eq:grpo_adv}) are computed over
all $G$ trajectories.
The pooled baseline therefore reflects the performance level achievable
under the scaffold, giving student rollouts a stronger calibrated reference
than a within-subset baseline would provide.
When $\Delta_e \le 0$, teacher-conditioned trajectories are excluded from
the RL loss entirely: advantages are computed over
$\mathcal{T}^{\mathsf{S}}$ alone, reducing the update to a standard GRPO
step.
Without this masking, non-beneficial teacher rollouts would still shift the
group baseline and distort student advantages---contaminating the policy
gradient even though distillation is gated off.

\subsection{Gain-Gated Privileged Distillation}\label{sec:distill}

The scaffolding in \S\ref{sec:scaffold} exposes the agent to successful
reasoning patterns it may not discover on its own, but because the retrieved
experience is unavailable at deployment, these gains remain ephemeral unless
consolidated into the policy itself.
This motivates a complementary mechanism: distilling the scaffold-induced
improvements into the base policy parameters so that the agent can reproduce
them from the standard context alone.

Our approach realizes this through asymmetric self-distillation.
Rather than relying on a separate, unconditionally superior teacher, we use
the same policy $\pi_\theta$ under two informational conditions: the teacher
is $\pi_\theta$ evaluated with the privileged context $c^{\mathsf{T}}$
(gradients stopped), while the student operates under $c^{\mathsf{S}}$.
The asymmetry therefore lies in the information available to each role, not
in model capacity.
The gain gate $\mathbb{I}(\Delta_e > 0)$ further ensures that distillation
is triggered only when the scaffold yields a verified performance advantage,
allowing the policy to selectively internalize reusable reasoning patterns
while avoiding negative transfer.

To guard against covariate shift, we construct the distillation target on
the student's own empirical support.
For a gain-gated task instance ($\Delta_e > 0$) and a student trajectory
$\tau \in \mathcal{T}^{\mathsf{S}}$ with token sequence
$(y_1, \ldots, y_m)$ generated under $c^{\mathsf{S}}$, we perform a
no-gradient forward pass of $\pi_\theta$ over the same tokens
conditioned on $c^{\mathsf{T}}$.
Because both passes share identical actions, the comparison isolates the
informational advantage of $e$ without introducing out-of-distribution
transitions.

We adopt the reverse KL divergence
$D_{\mathrm{KL}}(\pi_{\text{student}} \| \pi_{\text{teacher}})$ as the
distillation objective.
Unlike the forward KL, which compels the student to cover the full teacher
support and can induce mode-covering artifacts, the reverse KL is
mode-seeking: it encourages the policy to concentrate on the most effective
reasoning mode under the scaffold.
The token-level scaffold internalization loss is:
\vspace{-2mm}
\begin{align}\label{eq:distill}
  \mathcal{L}_{\text{distill}}(\theta)
  &= \mathbb{E}_{\tau \sim \mathcal{T}^{\mathsf{S}}} \!\Bigg[
    \mathbb{I}(\Delta_e\!>\!0) \nonumber \\[-2pt]
  & \qquad \sum_{t=1}^{m} \sum_{y \in \mathcal{V}}
    \pi_\theta^{\mathsf{S}}(y)\,\delta_t(y)
  \Bigg], \nonumber \\[4pt]
  \delta_t(y) &= \log \frac{\pi_\theta(y \mid c^{\mathsf{S}}_{<t})}
      {\pi_{\mathrm{sg}(\theta)}(y \mid c^{\mathsf{T}}_{<t})},
\end{align}
where $\pi_\theta^{\mathsf{S}}(y) \triangleq
\pi_\theta(y \mid c^{\mathsf{S}}_{<t})$ for brevity,
$\delta_t(y)$ is the per-token log-ratio between the student and the
privileged teacher, and $\mathrm{sg}(\cdot)$ denotes stop-gradient.
This loss is combined with the GRPO objective (Eq.~\eqref{eq:grpo}) to form
the joint actor loss:
\begin{equation}\label{eq:joint}
  \mathcal{L}_{\text{actor}}(\theta)
  = \mathcal{L}_{\text{RL}}(\theta)
  + \lambda\, \mathcal{L}_{\text{distill}}(\theta),
\end{equation}
where $\lambda$ controls the relative weight of scaffold internalization
versus the RL signal.
Through this joint optimization, the deployed agent reproduces privileged
reasoning without any external scaffold at test time.

\subsection{Experience Bank Evolution and Management}\label{sec:bank}

A static experience library cannot adequately support the agent throughout
training: as $\pi_\theta$ improves, it encounters situations where existing
experiences provide insufficient guidance; conversely, previously beneficial
experiences may become redundant once the policy has internalized the
corresponding behavior, or harmful if they conflict with newly discovered
strategies.
EDGE therefore treats $\mathcal{E}$ as a living repository that co-evolves
with the policy through both expansion and pruning.

Following \citet{xia2026skillrl}, we generate new experiences from the
agent's own rollout trajectories.
After each training step, we identify task categories whose
success rate falls below a threshold $\xi$ and collect
representative failed and successful trajectories from these
categories.
A reflector LLM then analyzes the contrast between them to synthesize new experiential guidance:
\begin{equation}\label{eq:reflect}
  e_{\text{new}}
  = f_{\text{reflect}}\!\left(\tau^{+},\, \tau^{-}\right),
\end{equation}
where $\tau^{+}$ and $\tau^{-}$ denote a successful and a failed trajectory
from the same task category, respectively.
The generated experiences are deduplicated against existing entries and
inserted into $\mathcal{E}$ for retrieval in subsequent training iterations.

Meanwhile, the utility of existing experiences is continuously tracked via
an Exponential Moving Average (EMA) score for each experience $e$:
\begin{equation}\label{eq:ema}
  U_e^{(t)} = (1 - \mu)\, U_e^{(t-1)} + \mu\, \Delta_e,
\end{equation}
where $\mu \in (0,1)$ is the momentum coefficient and $\Delta_e$ is the
instantaneous marginal gain from Eq.~\eqref{eq:gain}.
The EMA smooths over stochastic fluctuations while remaining responsive to
genuine shifts in experience utility.
Experiences whose tracked utility $U_e^{(t)}$ falls below a threshold $\eta$
are removed from $\mathcal{E}$, retiring scaffolds that have been absorbed
into the policy and reducing overhead during retrieval and rollout.
This yields a co-evolutionary dynamic: the policy improves by internalizing
useful experiences, exposing new failure modes that drive bank expansion,
while obsolete experiences are simultaneously pruned away.

\begin{table*}[t]
\centering
\resizebox{\textwidth}{!}{
\begin{tabular}{llccccccc|cc}
\toprule
\multirow{2}{*}{Type} & \multirow{2}{*}{Method} & \multicolumn{7}{c|}{\textbf{ALFWorld}} & \multicolumn{2}{c}{\textbf{WebShop}} \\
 & & Pick & Look & Clean & Heat & Cool & Pick2 & All & Score & Succ.\\
\midrule
\multicolumn{10}{l}{\textit{Closed-Source Model}} \\
Prompting& GPT-4o & 75.3 & 60.8 & 31.2 & 56.7 & 21.6 & 49.8 & 48.0& 31.8 & 23.7\\
Prompting& Gemini-2.5-Pro & 92.8 & 63.3 & 62.1 & 69.0 & 26.6 & 58.7 & 60.3& 42.5 & 35.9\\
\midrule
\multicolumn{10}{l}{\textit{Qwen2.5-1.5B-Instruct}} \\
Prompting& Base Model & 5.9 & 5.5 & 3.3 & 9.7 & 4.2 & 0.0 & 4.1 & 23.1 & 5.2\\
Prompting& ReAct & 17.4 & 20.5 & 15.7 & 6.2 & 7.7 & 2.0 & 12.8& 40.1& 11.3\\
Prompting& Reflexion & 35.3 & 22.2 & 21.7 & 13.6 & 19.4 & 3.7 & 21.8 & 55.8& 21.9\\
Post-Training& GRPO & 85.3 & 53.7 & \textbf{84.5} & 78.2 & 59.7 & 53.5 & 72.8& 75.8 & 56.8\\
Post-Training& SkillRL & \textbf{91.2}\textsubscript{\textpm4.3} & 64.3\textsubscript{\textpm4.6} & 78.1\textsubscript{\textpm5.4} & 76.9\textsubscript{\textpm6.3} & 70.8\textsubscript{\textpm6.5} & 54.6\textsubscript{\textpm6.1} & 74.2\textsubscript{\textpm4.7} & 77.3\textsubscript{\textpm3.5} & 60.9\textsubscript{\textpm3.7} \\
Post-Training& EMPO$^2$ & 86.9\textsubscript{\textpm3.3} & 66.2\textsubscript{\textpm5.2} & 79.3\textsubscript{\textpm4.9} & 79.1\textsubscript{\textpm4.5} & 75.3\textsubscript{\textpm5.7} & 64.8\textsubscript{\textpm6.6} & 76.8\textsubscript{\textpm3.7} & 78.2\textsubscript{\textpm3.7} & 63.2\textsubscript{\textpm3.3} \\
\rowcolor{gray!15}Post-Training& \textbf{\methodname{}} & 80.6\textsubscript{\textpm2.2} & \textbf{73.7}\textsubscript{\textpm5.5} & 76.0\textsubscript{\textpm4.3} & \textbf{87.3}\textsubscript{\textpm4.0} & \textbf{85.6}\textsubscript{\textpm6.1} & \textbf{68.4}\textsubscript{\textpm5.6} & \textbf{79.7}\textsubscript{\textpm2.3} & \textbf{78.8}\textsubscript{\textpm2.1} & \textbf{65.6}\textsubscript{\textpm3.2}\\
\midrule
\multicolumn{10}{l}{\textit{Qwen2.5-7B-Instruct}} \\
Prompting& Base Model & 33.4 & 21.6 & 19.3 & 6.9 & 2.8 & 3.2 & 14.8 & 26.4 & 7.8\\
Prompting& ReAct & 48.5 & 35.4 & 34.3 & 13.2 & 18.2 & 17.6 & 31.2 & 46.2 & 19.5\\
Prompting& Reflexion & 62.0 & 41.6 & 44.9 & 30.9 & 36.3 & 23.8 & 42.7& 58.1& 28.8\\
Post-Training& EvolveR\textsuperscript{\textdagger} & 64.9 & 33.3 & 46.4 & 13.3 & 33.3 & 33.3 & 43.8 & 42.5 & 17.6 \\
Post-Training& GRPO+Mem0\textsuperscript{\textdagger} & 78.1 & 54.8 & 56.1 & 31.0 & 65.0 & 26.9 & 54.7 & 58.1 & 37.5 \\
Post-Training& OPSD\textsuperscript{\textdagger} & 50.0 & 60.0 & 22.7 & 21.4 & 17.6 & 9.5 & 32.8& 4.5 & 2.3\\
Post-Training& GRPO & 92.8 & 85.7 & 89.3 & 75.7 & 74.5 & 67.7 & 82.1 & 80.3 & 70.1\\ 
Post-Training& GRPO+OPSD\textsuperscript{\textdagger} & 91.4 & 61.5 & \textbf{100} & 87.5 & 76.5 & 52.2 & 80.4& 86.8 & 76.5\\
Post-Training& SkillRL & 94.1\textsubscript{\textpm2.4} & 83.3\textsubscript{\textpm4.3} & 88.4\textsubscript{\textpm3.6} & 85.6\textsubscript{\textpm5.1} & 90.2\textsubscript{\textpm5.6} & 78.6\textsubscript{\textpm4.9} & 88.2\textsubscript{\textpm3.7} & 86.2\textsubscript{\textpm3.1} & 76.7\textsubscript{\textpm3.3} \\
Post-Training& EMPO$^2$ & 93.3\textsubscript{\textpm3.7} & \textbf{88.9}\textsubscript{\textpm3.9} & 91.2\textsubscript{\textpm4.4} & 88.5\textsubscript{\textpm5.3} & 89.4\textsubscript{\textpm4.6} & 79.8\textsubscript{\textpm5.1} & 89.1\textsubscript{\textpm2.8} & 88.3\textsubscript{\textpm2.6} & 77.1\textsubscript{\textpm4.0} \\
\rowcolor{gray!15}Post-Training& \textbf{\methodname{}} & \textbf{96.0}\textsubscript{\textpm1.9} & 
85.1\textsubscript{\textpm4.8}& 
93.4\textsubscript{\textpm3.2} & \textbf{90.0}\textsubscript{\textpm2.6} & \textbf{92.9}\textsubscript{\textpm4.5} & \textbf{81.6}\textsubscript{\textpm6.0} & \textbf{90.4}\textsubscript{\textpm2.2} & \textbf{89.6}\textsubscript{\textpm2.8} & \textbf{82.6}\textsubscript{\textpm3.8} \\
\bottomrule
\end{tabular}
}
\caption{Performance on ALFWorld and WebShop. We report the average success rate (\%) per subtask and overall for ALFWorld, and both the average score and success rate (\%) for WebShop. Results of \methodname{} and other experience-augmented training methods are averaged over 3 random seeds (mean$\pm$std). $\dagger$ denotes results replicated from \cite{xia2026skillrl} and \cite{lu2026selfdistilledagenticreinforcementlearning}.}
\vspace{-1mm}
\label{tab:main}
\end{table*}

\section{Experiments}

We evaluate \methodname{} to examine whether experience scaffolding can
improve agentic RL while being progressively internalized by the policy.
Our experiments address five questions:
(1)~Does \methodname{} improve over prompting, vanilla RL, and
prior experience-augmented methods, particularly on
exploration-intensive tasks?
(2)~Does the trained policy retain its performance when external
experiences are removed at inference time?
(3)~How much does each component---gain gating, distillation,
and pruning---contribute, and how do they interact?
(4)~Is experience utility truly non-stationary, and does the
co-evolutionary bank adapt accordingly during training?
(5)~Does \methodname{} generalize to a newer backbone and search-based QA,
and remain effective without a stronger proprietary reflector?

\subsection{Experiment Setup}
\paragraph{Environments.}
We evaluate \methodname{} across two complementary agentic settings:
interactive decision making and search-based QA. For the former, we use two
environments with sparse outcome-level feedback.
\textit{ALFWorld}~\citep{DBLP:conf/iclr/ShridharYCBTH21} comprises 3,827
text-based household tasks across six types (Pick, Look, Clean, Heat, Cool,
and Pick2), while \textit{WebShop}~\citep{DBLP:conf/nips/Yao0YN22} requires
agents to search and purchase products matching user instructions from a
catalog of over 1.1M items. To test generalization beyond interactive
environments, we further evaluate search-based QA on NQ, TriviaQA, PopQA,
HotpotQA, 2WikiMultiHopQA, MuSiQue, and Bamboogle, covering single- and
multi-hop evidence retrieval and reasoning.

\paragraph{Baselines.}
We compare against three families of methods: closed-source LLM agents (GPT-4o~\citep{openai2024gpt4ocard} and Gemini-2.5-Pro~\citep{comanici2025gemini25}), prompting agents (ReAct~\citep{yao2023react} and Reflexion~\citep{shinn2024reflexion}), and training-based agents.
The post-training baselines include GRPO~\citep{shao2024deepseekmath}, OPSD~\citep{zhao2026selfdistilledreasoneronpolicyselfdistillation}, EvolveR~\citep{wu2026evolver}, GRPO augmented with Mem0~\citep{chhikara2025mem0}, SkillRL~\citep{xia2026skillrl}, and EMPO$^2$~\citep{liu2026exploratorymemoryaugmentedllmagent}.
Detailed descriptions are provided in Appendix~\ref{appendix:baseline_detail}.

\paragraph{Training details.}
We use Qwen2.5-1.5B/7B-Instruct~\citep{qwen2025qwen25technicalreport} as the base models for ALFWorld and WebShop, and Qwen3-4B-Instruct-2507~\citep{yang2025qwen3technicalreport} for search-based QA. For ALFWorld and WebShop, all post-training methods use exactly the same hyperparameter configurations. The rollout group size $G$ for group-based RL methods is set to 8. For experience retrieval, we encode experiences with Qwen3-Embedding-0.6B~\citep{zhang2025qwen3embeddingadvancingtext} and rank candidates by cosine similarity against the task instruction and initial observations. We use GPT-4o~\citep{openai2024gpt4ocard} as the default reflector. To test whether EDGE depends on a stronger proprietary model, the QA experiments also replace GPT-4o with the Qwen3 policy itself while leaving the remaining framework unchanged. Full training settings and hyperparameter details are provided in Appendix~\ref{appendix:train_detail}, with a wall-clock breakdown in Appendix~\ref{appendix:compute_cost}.

\subsection{Main Results}\label{sec:main_results}
Table~\ref{tab:main} presents results on ALFWorld and WebShop across two
model scales.
At 7B, \methodname{} achieves 90.4\% success rate on ALFWorld and 82.6\% on WebShop, improving over GRPO by 8.3 and 12.5 points and over the strongest
experience-augmented baseline EMPO$^2$ by 1.3 and 5.5 points,
with consistent gains at 1.5B.

The subtask breakdown reveals that \methodname{}'s advantage concentrates
on exploration-heavy tasks---Heat, Cool, and Pick2---which demand longer
action sequences with fewer intermediate rewards.
On WebShop, the success-rate improvement over GRPO (+12.5) exceeds the
score improvement (+9.3), indicating that \methodname{} helps agents
complete full decision chains rather than merely accumulate partial
credit.

These results highlight that experience is most effective as selective
exploration guidance rather than unconditional augmentation.
Naive memory approaches (EvolveR, GRPO+Mem0) degrade well below vanilla
GRPO, and standalone self-distillation (OPSD) provides insufficient
signal without effective exploration.
Combining the two (GRPO+OPSD) recovers competitive aggregate performance
but remains brittle across subtasks, whereas \methodname{}'s
marginal-gain gating ensures only beneficial experiences contribute,
yielding both higher overall success and more uniform subtask coverage.

\begin{table*}[t]
\centering
\setlength{\tabcolsep}{6pt}
\renewcommand{\arraystretch}{1.05}
\resizebox{\textwidth}{!}{
\begin{tabular}{lcccccccc}
\toprule
Method & NQ & TriviaQA & PopQA & HotpotQA & 2Wiki & MuSiQue & Bamboogle & Avg. \\
\midrule
Qwen3-4B-Instruct-2507 & 17.8 & 42.3 & 27.2 & 23.6 & 21.5 & 5.6 & 7.4 & 20.8 \\
Search-R1 & 40.6 & 61.2 & 43.7 & 35.2 & 36.3 & 12.4 & 41.6 & 38.7 \\
SkillRL (GPT-4o) & 42.4 & 59.6 & 44.3 & 42.1 & 43.4 & 15.9 & 43.2 & 41.5 \\
\rowcolor{gray!15}EDGE (GPT-4o) & \textbf{48.0} & 63.4 & \textbf{46.9} & 42.4 & \textbf{48.1} & \textbf{17.8} & \textbf{45.6} & \textbf{44.6} \\
\rowcolor{gray!15}EDGE (self) & 44.8 & \textbf{64.3} & 45.4 & \textbf{43.6} & 45.5 & 16.2 & 44.2 & 43.4 \\
\bottomrule
\end{tabular}}
\caption{Performance (\%) on seven search-based QA benchmarks with Qwen3-4B-Instruct-2507. ``GPT-4o'' and ``self'' denote the reflector used to synthesize experiences. Best results are bolded.}
\vspace{-2mm}
\label{tab:qwen3-qa}
\end{table*}

\paragraph{Generalization to Search-Based QA and Self-Reflection.}
Table~\ref{tab:qwen3-qa} tests whether EDGE transfers to a newer backbone and search-based QA without relying on external reflection. With GPT-4o, EDGE leads on five of seven benchmarks. More importantly, self-reflection retains 97.3\% of this performance while surpassing both baselines on every benchmark. Its particularly strong results on TriviaQA and HotpotQA further suggest that self-generated experience remains effective across both fact-oriented and multi-hop search settings. Together, these findings indicate that EDGE's gains arise primarily from its experience construction mechanism rather than privileged access to a stronger reflector; GPT-4o improves the quality of the resulting experience, but is not essential for successful transfer.

\begin{figure}[t]
\centering
    \includegraphics[width=\linewidth]{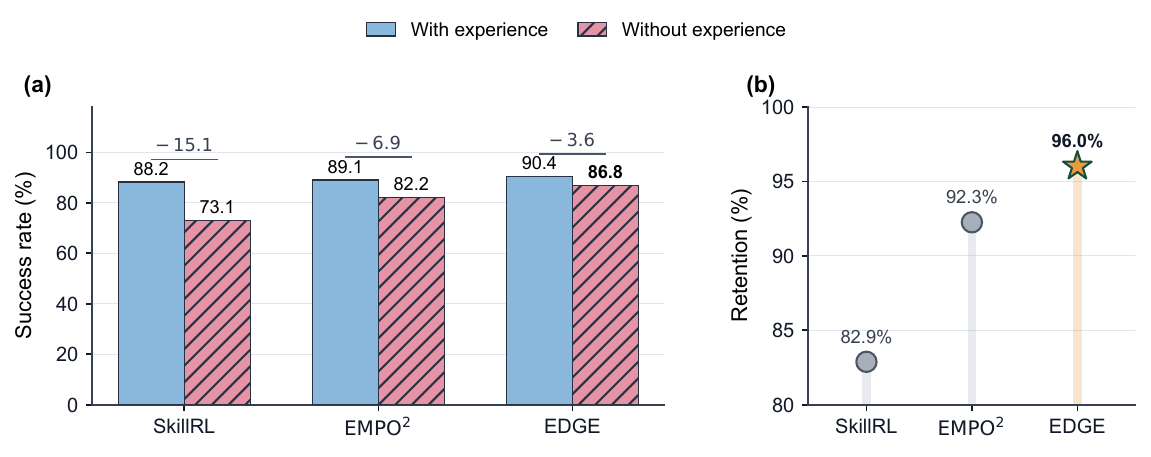}
\caption{
\textbf{Performance retention after scaffold removal at inference time.}
\methodname{} preserves 96.0\% of its scaffolded performance without
external experiences, compared with 82.9\% for SkillRL and 92.3\% for
EMPO$^2$, indicating more effective internalization into the parametric policy at the 7B scale.
}
\vspace{-2mm}
\label{fig:scaffold}
\end{figure}

\begin{figure*}[t]
\centering
    \includegraphics[width=0.99\textwidth]{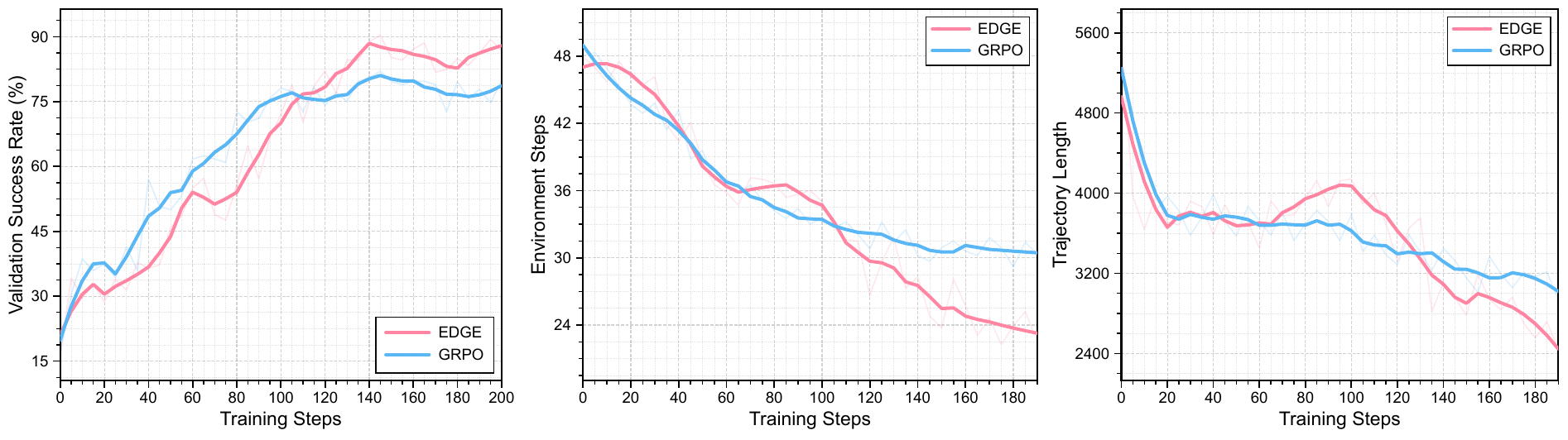}
\caption{
\textbf{Training dynamics of \methodname{} vs.\ GRPO on ALFWorld with Qwen2.5-7B-Instruct.}
\methodname{} achieves higher validation success while reducing both
environment steps and trajectory length more rapidly, indicating that
scaffolded exploration is progressively internalized into more efficient
experience-free behavior.
}

\label{fig:training-dynamics}
\end{figure*}

\paragraph{Inference without External Scaffolds.}
Figure~\ref{fig:scaffold} poses a stricter test: how much performance
survives when all external experiences are withheld at inference time?
\methodname{} preserves 96.0\% of its scaffolded performance, compared
with 92.3\% for EMPO$^2$ and 82.9\% for SkillRL, confirming that the
reverse-KL distillation stage successfully transfers scaffold-induced
behavior into the parametric policy.
This near-complete
retention supports the intended scaffold-to-parameter transition:
experience-guided behaviors are largely internalized by the policy and
remain available without inference-time retrieval.

\subsection{Analysis}\label{sec:ablation}
\paragraph{Ablation Studies.}

\begin{table}[t]
\centering

\resizebox{\linewidth}{!}{
\begin{tabular}{lccccccc}
\toprule
Method & Pick & Look & Clean & Heat & Cool & Pick2 & All \\
\midrule
GRPO & 92.8 & \textbf{85.7} & 89.3 & 75.7 & 74.5 & 67.7 & 82.1 \\
\midrule
\rowcolor{gray!15}
\textbf{\methodname{}} & \textbf{96.0} & 85.1 & \textbf{93.4} & \textbf{90.0} & \textbf{92.9} & \textbf{81.6} & \textbf{90.4} \\
w/o ExpPruning & 90.5 & 82.2 & 90.1 & 84.3 & 87.6 & 76.8 & 86.7 \\
w/o Distillation & 91.2 & 81.5 & 91.8 & 78.7 & 77.2 & 71.1 & 83.6 \\
w/o Gain-Gating & 83.1 & 72.5 & 79.7 & 65.3 & 61.8 & 61.2 & 72.3 \\
\bottomrule
\end{tabular}}
\caption{\textbf{Ablation results} on ALFWorld with Qwen2.5-7B-Instruct. We report success rate (\%) per subtask and overall.}
\vspace{-2mm}
\label{tab:ablation}
\end{table}

Table~\ref{tab:ablation} isolates each component on ALFWorld
with Qwen2.5-7B-Instruct.
The most striking finding is that removing the gain gate
drops overall success to 72.3\%---9.8 points below
vanilla GRPO---revealing that unfiltered experience injection
does not merely fail to help but actively harms the policy,
particularly on exploration-heavy subtasks where misleading
guidance compounds over long horizons.
This result also addresses a natural concern about the
$G/2{+}G/2$ rollout partition: since all other ablated
variants retain the same split yet outperform GRPO, the
partition itself does not dilute the RL signal; the
degradation is attributable entirely to distilling
low-quality experiences.

The remaining two components contribute complementary
benefits.
Without distillation, performance falls to 83.6\%, only
1.5 points above GRPO, with losses concentrated on the
same exploration-heavy subtasks---confirming that
scaffolded exploration provides transient guidance but
does not durably reshape the policy without explicit
behavioral transfer.
Without experience pruning, performance declines more
modestly to 86.7\% with losses spread evenly across
subtasks, indicating that pruning acts as a maintenance
mechanism that keeps the experience bank aligned with
the policy's evolving capability rather than targeting
any specific failure mode.

\paragraph{Training Dynamics.}
Figure~\ref{fig:training-dynamics} reveals a clear divergence between
EDGE and GRPO after approximately step~100: GRPO plateaus and begins to
regress, whereas EDGE continues to improve steadily toward 90\%
validation success.
We attribute GRPO's decline to an exploration--exploitation collapse:
once the policy commits to locally successful strategies, it loses the
diversity needed to solve the remaining hard tasks, and further
optimization erodes earlier gains.
EDGE's experience scaffolds counteract this by continually injecting
structured exploration guidance, while gain gating ensures that this
guidance remains beneficial as the policy strengthens.
The efficiency panels corroborate this interpretation---EDGE reduces
both environment steps and trajectory length more rapidly than GRPO,
indicating that the policy internalizes increasingly direct action
sequences rather than relying on extended trial-and-error, consistent
with the scaffold-removal results in Figure~\ref{fig:scaffold}.

\begin{figure}[t]
\centering
    \includegraphics[width=\linewidth]{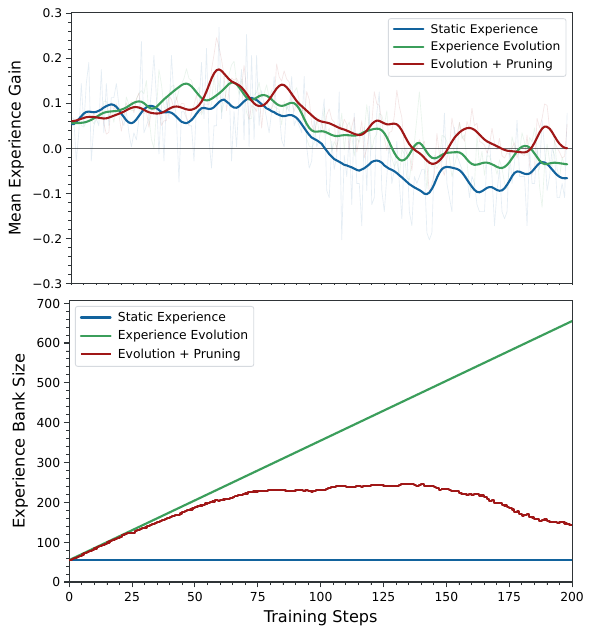}
\caption{Experience gain dynamics during training.}
\label{fig:gain-tracking}
\vspace{-2mm}
\end{figure} 

\paragraph{Experience Gain Tracking.}
Figure~\ref{fig:gain-tracking} tracks the mean marginal gain
$\Delta_e$ (Eq.~\eqref{eq:gain}) across training to test a key
premise of \S\ref{sec:bank}: that experience utility is
non-stationary.
With a static bank, the initially positive gain decays and turns
negative after roughly step~100---confirming that once-useful
experiences become actively harmful as the policy outgrows them,
and explaining why removing gain gating in
Table~\ref{tab:ablation} degrades performance below vanilla GRPO.
Experience evolution delays this decay, but unchecked bank growth
(reaching over 650 entries) introduces retrieval noise that keeps
the gain signal volatile.
The full co-evolutionary configuration sustains the most stable
positive gain: pruning not only curbs bank growth but causes the
bank to shrink after step~100, indicating that the policy
absorbs existing experiences faster than new failure modes
generate replacements---a direct signature of successful
internalization.
Further details on bank evolution dynamics appear in
Appendix~\ref{appendix:bank_evolution}.

\section{Related Work}\label{sec:related}

\paragraph{Reinforcement Learning for LLM Agents.}
RL has become a standard post-training paradigm for LLM agents in
multi-turn environments~\citep{DBLP:conf/nips/Yao0YN22,DBLP:conf/iclr/ShridharYCBTH21,feng2025gigpo,wang2025ragen,jin2025searchr1,li-etal-2026-whats},
progressing from PPO-based methods~\citep{schulman2017ppo} to
critic-free objectives such as GRPO~\citep{shao2024deepseekmath}
and RLOO~\citep{DBLP:conf/acl/AhmadianCGFKPUH24}, with further
improvements in multi-turn credit assignment through turn-level
or stepwise
signals~\citep{feng2025gigpo,wei2025reinforcingmultiturnreasoningllm,wang2025sparlreinforcingllmagents,xie-etal-2026-unlocking,zhang2026rapoexpandingexplorationllm}.
However, the reusable exploration patterns within trajectories are
still consumed once and discarded.
\methodname{} retains the group-sampling backbone of GRPO but
repurposes it to estimate which retrieved experiences currently
improve exploration and to transfer their effects into the policy.

\paragraph{Experience-Augmented LLM Agents.}
External memory and experience reuse have been explored through
prompting-based reflections~\citep{shinn2024reflexion,zhao2024expel,yang2024bufferthought,fang2026mempexploringagentprocedural,ouyang2026reasoningbank}
and persistent retrieval during
interaction~\citep{chhikara2025mem0,xia2026skillrl,wu2026evolver,ma2026freshnessaware,wang2026reinforcementlearningselfimprovingagent,zhang2026memskilllearningevolvingmemory}.
Recent methods selectively replay past reasoning
traces~\citep{yan2025learningreasonoffpolicyguidance,zhang2025stephintmultilevelstepwisehints,qin2025learnropestrustwins,zhan2026exgrpo},
but primarily target single-turn tasks without the multi-turn,
partially observable interaction loops of agentic settings.
EMPO$^2$~\citep{liu2026exploratorymemoryaugmentedllmagent} pursues
memory-free behavior through heuristic experience selection and off-policy
distillation. Concurrent SKILL0~\citep{lu2026skill0incontextagenticreinforcement}
progressively withdraws a preconstructed skill inventory under a decaying
curriculum. In contrast, \methodname{} uses paired-rollout gains to validate
and distill online experience while co-evolving its bank through utility-based
pruning.

\paragraph{Privileged Information and Self-Distillation.}
Leveraging training-time information unavailable at deployment spans
LUPI~\citep{vapnik2009lupi}, asymmetric actor-critic
methods~\citep{pinto2017asymmetricactorcriticimagebased}, and context
distillation in
LLMs~\citep{snell2022learningdistillingcontext,choudhury2025leap,liu2026exploratorymemoryaugmentedllmagent,cheng2026mem2evolveselfevolvingagentscoevolutionary},
though these approaches are largely off-policy and suffer from distribution
mismatch with the student's own visitation.
On-policy
distillation~\citep{agarwal2024onpolicydistillation,ye2026onpolicycontextdistillationlanguage}
addresses this by supervising the student on its own sequences, and
on-policy self-distillation
(OPSD)~\citep{zhao2026selfdistilledreasoneronpolicyselfdistillation,hubotter2026rlsd,lu2026selfdistilledagenticreinforcementlearning}
further removes the need for a separate teacher.
\methodname{} shares this structure but adds gain-based distillation gating to activate
distillation only under verified positive gain and co-evolves the
experience bank through utility-driven expansion and pruning.


\section{Conclusion}

We presented \methodname{}, a framework for using retrieved experience as a temporary training-time scaffold rather than a persistent inference-time dependency.
EDGE estimates the marginal utility of retrieved experiences under the current policy, admits only positive-gain scaffolds, and distills their behavioral effect into the standard experience-free policy.
Across ALFWorld, WebShop, and seven search-based QA benchmarks, EDGE improves over RL and experience-augmented baselines, with especially large gains on exploration-intensive subtasks and strong retention after scaffold removal.
The Qwen3 results further show that these gains transfer to a newer backbone and largely persist when the policy itself replaces GPT-4o as the reflector.
These results suggest a practical principle for agentic RL: external experience is most useful when it is dynamically validated, selectively applied, and ultimately internalized.

\section*{Limitations}

While \methodname{} requires no extra environment rollouts,
it introduces training-time overhead for maintaining the
experience bank, computing teacher--student comparisons, and
invoking the reflector LLM.
The effectiveness of gain-gating also depends on reward
quality; as with all outcome-based RL methods, noisy or
misspecified rewards would reduce the reliability of the
marginal-gain estimates.
In terms of empirical scope, our evaluation covers two interactive
environments and seven search-based QA benchmarks with Qwen2.5 models at
the 1.5B and 7B scales and Qwen3 at the 4B scale; broader validation across
model families, larger scales, and multimodal or real-world environments
remains future work.
More broadly, \methodname{} assumes experiences are discrete
textual artifacts; extending the scaffold-to-parameter
principle to latent memory representations is an open problem.

\section*{Acknowledgements}
This work is supported by the Strategic Priority Research Program of Chinese Academy of Sciences under Grant XDA04080400 and Beijing Natural Science Foundation L259016.

\bibliography{custom}

\clearpage
\appendix
\section{Theoretical Analysis of \methodname{}}
\label{app:theory}

This section provides an analytical justification for the four core
components of \methodname{}, following the pipeline through which a
retrieved experience~$e$ influences the policy update:
its marginal utility is estimated (\S\ref{subsec:theory_vopi}),
smoothed across training steps (\S\ref{subsec:theory_ema}),
used to gate and calibrate the RL advantage
(\S\ref{subsec:theory_advantage}), and channeled through on-support
distillation (\S\ref{subsec:theory_kl}).

Throughout, let $c^{\mathsf{S}}$ denote the standard context,
$c^{\mathsf{T}}=c^{\mathsf{S}}\oplus e$ the privileged context,
$\pi_\theta$ the current policy, and $R(\tau)\in\{0,1\}$ the binary outcome
reward. Teacher and student rollout subsets
$\mathcal{T}^{\mathsf{T}},\mathcal{T}^{\mathsf{S}}$ each have size
$K=G/2$. All expectations and probabilities condition on fixed
$\pi_\theta$ and~$e$.

\subsection{Marginal-Gain Estimation and Gating}
\label{subsec:theory_vopi}

Define the \emph{value of privileged information} as the expected return
gap between the privileged and standard contexts:
\begin{equation}
\label{eq:vopi}
  \mathcal{V}_e(\theta)
  \;=\;
  \mathbb{E}_{\tau\sim\pi_\theta(\cdot\mid c^{\mathsf{T}})}
  \!\bigl[R(\tau)\bigr]
  \;-\;
  \mathbb{E}_{\tau\sim\pi_\theta(\cdot\mid c^{\mathsf{S}})}
  \!\bigl[R(\tau)\bigr].
\end{equation}
The empirical marginal gain is
\begin{equation}
\label{eq:gain_empirical}
  \Delta_e
  \;=\;
  \frac{1}{K}\!\sum_{\tau\in\mathcal{T}^{\mathsf{T}}}\!R(\tau)
  \;-\;
  \frac{1}{K}\!\sum_{\tau\in\mathcal{T}^{\mathsf{S}}}\!R(\tau).
\end{equation}
Under the assumption that the two subsets are conditionally independent and
identically distributed up to the presence of~$e$, $\Delta_e$ is unbiased
for $\mathcal{V}_e(\theta)$ with variance
$(p_T(1{-}p_T)+p_S(1{-}p_S))/K$, where $p_T$ and $p_S$ are the respective
success probabilities. Decomposing $\Delta_e$ as a sum of $2K$ independent
terms each bounded in an interval of length $1/K$, Hoeffding's inequality
yields
\begin{equation}
\label{eq:gain_concentration}
  \Pr\!\bigl(|\Delta_e-\mathcal{V}_e(\theta)|\ge\epsilon\bigr)
  \;\le\;
  2\exp\!\bigl(-K\epsilon^2\bigr).
\end{equation}

\methodname{} converts this estimate into a binary gate:
\begin{equation}
\label{eq:gate}
  M_e \;=\; \mathbb{I}(\Delta_e > 0).
\end{equation}
The gate serves as a probabilistic risk-control mechanism. By the one-sided
form of~\eqref{eq:gain_concentration}, if an experience is truly harmful
with margin $\gamma>0$ ($\mathcal{V}_e(\theta)\le-\gamma$), then
$\Pr(M_e{=}1)\le\exp(-K\gamma^2)$; the symmetric bound holds for missed
activations when $\mathcal{V}_e(\theta)\ge\gamma$. When $M_e=0$, teacher
rollouts are masked from both the RL and distillation losses, and the update
falls back to standard GRPO on the student subset.

\subsection{EMA Smoothing}
\label{subsec:theory_ema}

Because $\Delta_e$ is high-variance for small $K$ and the true utility
$\mathcal{V}_e(\theta)$ drifts as the policy evolves, \methodname{} tracks
each experience's utility with an exponential moving average:
\begin{equation}
\label{eq:ema}
  U_e^{(t)}
  \;=\;
  (1-\mu)\,U_e^{(t-1)}
  \;+\;
  \mu\,\Delta_e^{(t)},
  \qquad \mu\in(0,1).
\end{equation}
Under a local-stationarity approximation (successive $\Delta_e^{(t)}$
treated as i.i.d.\ with variance $\sigma_e^2$), the steady-state variance
is $\mathrm{Var}(U_e)=\frac{\mu}{2-\mu}\,\sigma_e^2$, which is strictly
less than $\sigma_e^2$ for any $\mu<1$. Smaller $\mu$ yields greater
smoothing at the cost of slower adaptation to genuine utility shifts. The
pruning threshold $\eta$ thus operates on a smoothed estimate of recent
marginal utility rather than a single noisy contrast.

\subsection{Pooled Advantage Calibration}
\label{subsec:theory_advantage}

When the gate activates, teacher rollouts enter the RL update and alter the
advantage baseline. Let $b_S,b_T$ be the student and teacher mean returns,
so $\Delta_e=b_T-b_S$. The active set is
\begin{equation}
\label{eq:active_set}
  \mathcal{A}
  \;=\;
  \begin{cases}
  \mathcal{T}^{\mathsf{S}}\cup\mathcal{T}^{\mathsf{T}}, & M_e=1,\\[2pt]
  \mathcal{T}^{\mathsf{S}}, & M_e=0,
  \end{cases}
\end{equation}
with baseline $b_{\mathcal{A}}=\mathrm{mean}_{j\in\mathcal{A}}(R_j)$. When
$M_e=1$, the pooled baseline
$b_{\mathrm{pool}}=\frac{1}{2}(b_S+b_T)$ shifts the raw advantage
numerators (prior to GRPO's standard-deviation normalization, which rescales
uniformly without changing signs) as follows:
\begin{align}
\label{eq:advantage_shift_s}
  \widetilde{A}_S^{\methodname{}}
  &= \widetilde{A}_S^{\mathrm{within}}
     - \tfrac{1}{2}\Delta_e,
  \\[2pt]
\label{eq:advantage_shift_t}
  \widetilde{A}_T^{\methodname{}}
  &= \widetilde{A}_T^{\mathrm{within}}
     + \tfrac{1}{2}\Delta_e,
\end{align}
where the superscript ``within'' denotes baselines computed from each
subset alone. Since $M_e=1$ implies $\Delta_e>0$, student advantages
are shifted \emph{downward} and teacher advantages \emph{upward}:
privileged successes receive stronger reinforcement, while unguided
successes are tempered when the scaffold demonstrates superior performance.
When $M_e=0$, the teacher subset is excluded and no shift is applied.

\subsection{On-Support Reverse-KL Distillation}
\label{subsec:theory_kl}

Directly imitating teacher-generated trajectories risks covariate
shift~\citep{pmlr-v15-ross11a}, as the teacher may visit prefixes whose
rationale depends on information absent from the student context (causal
misidentification; \citealp{NEURIPS2019_94701864}). \methodname{} mitigates
this by evaluating the teacher only on student-generated prefixes. For a
student trajectory
$\tau=(y_1,\ldots,y_m)\in\mathcal{T}^{\mathsf{S}}$, let
$h_t^{\mathsf{S}}=(c^{\mathsf{S}},y_{<t})$ and
$h_t^{\mathsf{T}}=(c^{\mathsf{T}},y_{<t})$. The on-support distillation
loss is
\begin{equation}
\label{eq:distill}
\begin{split}
  \widehat{\mathcal{L}}_{\mathrm{distill}}(\theta)
  \;=\;\;
  &M_e
  \!\sum_{\tau\in\mathcal{T}^{\mathsf{S}}}
  \sum_{t=1}^{|\tau|}
  \\
  &D_{\mathrm{KL}}\!\bigl(
    \pi_\theta(\cdot\mid h_t^{\mathsf{S}})
    \;\big\|\;
    \pi_{\mathrm{sg}(\theta)}(\cdot\mid h_t^{\mathsf{T}})
  \bigr),
\end{split}
\end{equation}
where $\mathrm{sg}(\cdot)$ denotes stop-gradient. Unlike forward-KL
imitation on teacher rollouts, this objective compares distributions only at
prefixes the student has actually reached, avoiding training on
teacher-only states. The reverse-KL direction is mode-seeking with respect
to the teacher, concentrating student mass on teacher-preferred actions
rather than spreading to cover the full teacher distribution, and thereby
keeping the update conservative on the student support.

The final actor loss combines both pathways:
\begin{equation}
\label{eq:joint}
  \mathcal{L}_{\mathrm{actor}}(\theta)
  \;=\;
  \mathcal{L}_{\mathrm{RL}}(\theta;\mathcal{A})
  \;+\;
  \lambda\,\widehat{\mathcal{L}}_{\mathrm{distill}}(\theta).
\end{equation}
When $M_e=0$, both terms reduce to unprivileged-only updates
($\mathcal{A}=\mathcal{T}^{\mathsf{S}}$,
$\widehat{\mathcal{L}}_{\mathrm{distill}}=0$). When $M_e=1$, the teacher
subset raises the RL baseline (\S\ref{subsec:theory_advantage}) and the
reverse KL transfers privileged behavior into the standard-context policy.

\section{Implementation Details}
\subsection{Baselines}
\label{appendix:baseline_detail}
\begin{itemize}
    \item[\textbullet] \textbf{GPT-4o \citep{openai2024gpt4ocard}}: A closed-source multimodal LLM from OpenAI, used as a strong proprietary agent baseline with standard prompting.
    \item[\textbullet] \textbf{Gemini-2.5-Pro \citep{comanici2025gemini25}}: A closed-source reasoning model from Google DeepMind, serving as another proprietary agent baseline with standard prompting.
    \item[\textbullet] \textbf{ReAct \citep{yao2023react}}: A prompting framework that interleaves chain-of-thought reasoning with environment actions, enabling LLMs to plan and act in a synergistic loop.
    \item[\textbullet] \textbf{Reflexion \citep{shinn2024reflexion}}: Extends ReAct by appending verbal self-reflection after task failures, allowing the agent to refine its strategy across successive trials without weight updates.
    \item[\textbullet] \textbf{GRPO \citep{shao2024deepseekmath}}: A group-relative policy optimization algorithm that estimates advantages from a group of sampled rollouts, eliminating the need for a separate critic network.
    \item[\textbullet] \textbf{OPSD \citep{zhao2026selfdistilledreasoneronpolicyselfdistillation}}: An on-policy self-distillation method that distills the model's own high-quality rollouts back into itself to improve reasoning without external supervision.
    \item[\textbullet] \textbf{GRPO+OPSD}: A hybrid baseline that combines GRPO's group-relative advantage estimation with OPSD's on-policy self-distillation objective.
    \item[\textbullet] \textbf{EvolveR \citep{wu2026evolver}}: An experience-augmented training method that iteratively evolves a retrieval-augmented memory of past trajectories to guide policy learning.
    \item[\textbullet] \textbf{GRPO+Mem0 \citep{chhikara2025mem0}}: Augments GRPO with Mem0, a memory module that stores and retrieves past interaction experiences as additional context during both training and inference.
    \item[\textbullet] \textbf{Search-R1 \citep{jin2025searchr1}}: Trains language models with reinforcement learning to interleave reasoning and search, serving as the standard search-agent baseline in our QA evaluation.
    \item[\textbullet] \textbf{SkillRL \citep{xia2026skillrl}}: A skill-based RL framework that extracts reusable skills from successful trajectories and conditions policy optimization on retrieved skill demonstrations.
    \item[\textbullet] \textbf{EMPO$^2$ \citep{liu2026exploratorymemoryaugmentedllmagent}}: An exploratory memory-augmented policy optimization method that leverages curated past experiences to enhance exploration during RL training.
\end{itemize}
These baselines span closed-source LLM agents, prompting-based reasoning frameworks, standard RL algorithms, self-distillation methods, and experience-augmented training approaches, enabling a comprehensive evaluation of \methodname{} from multiple perspectives.

\subsection{RL-Training Configuration}
\label{appendix:train_detail}
We implement \methodname{} on top of the verl-agent framework~\citep{feng2025gigpo}
and train the model with the joint optimization objective in
Eq.~\ref{eq:joint}. To reduce the computational cost of reverse-KL
distillation, we approximate the vocabulary-level loss using only the
top-$k$ student tokens with the highest log probabilities. The experience bank is initialized as empty, and each newly inserted experience is assigned an initial utility score of 0. For retrieval, we use the task instruction and
initial environment observations as the query, encode experiences with
Qwen3-Embedding-0.6B~\citep{zhang2025qwen3embeddingadvancingtext}, and rank candidates by cosine similarity. We retrieve
a top-$m$ candidate pool and use the highest-scoring experience as the
scaffold for the current rollout.

After each training step, we update the experience bank through both
expansion and pruning. For expansion, if the success rate of a task category
falls below the expansion threshold $\xi$, the reflector contrasts successful
and failed trajectories from the same category and synthesizes up to three
new experiences. We use GPT-4o~\citep{openai2024gpt4ocard} by default; in the
EDGE (self) QA variant, the Qwen3 policy itself serves as the reflector, with
the remaining framework unchanged. For pruning, we update the EMA
utility score of each experience according to Eq.~\ref{eq:ema} and remove
experiences whose utility falls below the pruning threshold $\eta$. For both
ALFWorld and WebShop, we use the hyperparameters in
Table~\ref{train_hyperparameters}. All training and profiling experiments are
conducted on $8 \times$ A100 80GB GPUs.

\begin{table}[!htb]
\centering
\setlength{\tabcolsep}{4pt}
\renewcommand{\arraystretch}{1.08}
\begin{tabular*}{\linewidth}{@{\extracolsep{\fill}}ll@{}}
\toprule
\textbf{Configuration} & \textbf{Value} \\
\midrule
\multicolumn{2}{@{}l}{\textit{RL-Training}} \\
Actor learning rate & $1\text{e}^{-6}$ \\
Maximum prompt length & 4096 \\
Maximum response length & 512 \\
Training batch size & 16 \\
Rollout group size ($G$) & 8 \\
Training mini-batch size & 128  \\
Rollout temperature & 1.0 \\
Training steps & 200 \\
\midrule
\multicolumn{2}{@{}l}{\textit{EDGE configuration}} \\
Distillation weight $\lambda$ & 0.1 \\
Distillation top-$k$ tokens & 20 \\
Utility pruning threshold $\eta$ & -0.1 \\
Retrieval pool size (top-$m$) & 6 \\
Expansion success threshold $\xi$ & 0.4 \\
EMA momentum $\mu$ & 0.5 \\
Maximum new experiences per step & 3 \\
\bottomrule
\end{tabular*}
\vspace{-5pt}
\caption{RL Hyperparameters}\label{train_hyperparameters}
\end{table}

\subsection{Computational Cost}
\label{appendix:compute_cost}
We profile all methods under the same hardware and training configuration
using Qwen2.5-7B-Instruct on ALFWorld. Table~\ref{tab:compute_cost} reports
wall-clock seconds per training step. Rollout time includes experience
retrieval and prompt construction, while reflector latency is reported
separately.

\begin{table}[!htb]
\centering
\resizebox{\linewidth}{!}{
\begin{tabular}{lccccc}
\toprule
Method & Rollout & Old Prob. & Ref. Prob. & Update & Reflector \\
\midrule
GRPO & 263.41 & 14.10 & 14.23 & 55.38 & -- \\
SkillRL & 334.13 & 18.45 & 18.51 & 64.74 & 17.12 \\
EMPO$^2$ & 318.69 & 17.37 & 17.65 & 63.17 & 27.56 \\
\rowcolor{gray!15}\methodname{} & 304.56 & 17.26 & 17.67 & 61.45 & 16.78 \\
\bottomrule
\end{tabular}}
\caption{Per-step wall-clock time (seconds) on ALFWorld with
Qwen2.5-7B-Instruct.}
\label{tab:compute_cost}
\end{table}

Summing all components, \methodname{} requires 417.72 seconds per step, a
20.3\% overhead over GRPO, mainly from retrieval, privileged-context forward
computation, and reflection. Nevertheless, it is 7.8\% faster than SkillRL
and 6.0\% faster than EMPO$^2$. All methods use the same rollout budget;
\methodname{} partitions the original group into experience-conditioned and
experience-free trajectories without additional environment interactions.

\begin{table*}[t]
\centering
\small
\setlength{\tabcolsep}{5pt}
\renewcommand{\arraystretch}{1.12}
\begin{tabularx}{\textwidth}{
  >{\raggedright\arraybackslash}p{0.13\textwidth}
  >{\raggedright\arraybackslash}X
  >{\raggedright\arraybackslash}X
  >{\raggedright\arraybackslash}X
}
\toprule
Case & Failure mode & Reflected experience & Later evidence \\
\midrule
1. Clean bowl &
Navigates to destination before finding object &
Locate object before placing &
Retrieved on a related task at step 155; $U{=}0.125$ \\
2. Pillow on sofa &
Issues \texttt{take} from source lacking target &
Verify source before taking &
Utility rises to $0.578$ by step 200 (78 retrievals) \\
3. Pick2 search &
Unstructured search before collecting targets &
Search systematically before storing &
Retrieved on a related Pick2 task at step 166 \\
4. No-op movement &
Repeats navigation after arrival &
Avoid repeated no-op moves &
Near-threshold utility; recovers to $U{=}0.452$ by step 200 \\
\bottomrule
\end{tabularx}
\caption{\textbf{Experience-bank co-evolution examples.} Each case is
extracted from saved failure trajectories, LLM reflection logs, retrieval
logs, and utility traces.}
\label{tab:bank_case_summary}
\end{table*}

\section{Further Analysis}
\label{appendix:further_analysis}

\begin{figure}[t]
\centering
    \includegraphics[width=\linewidth]{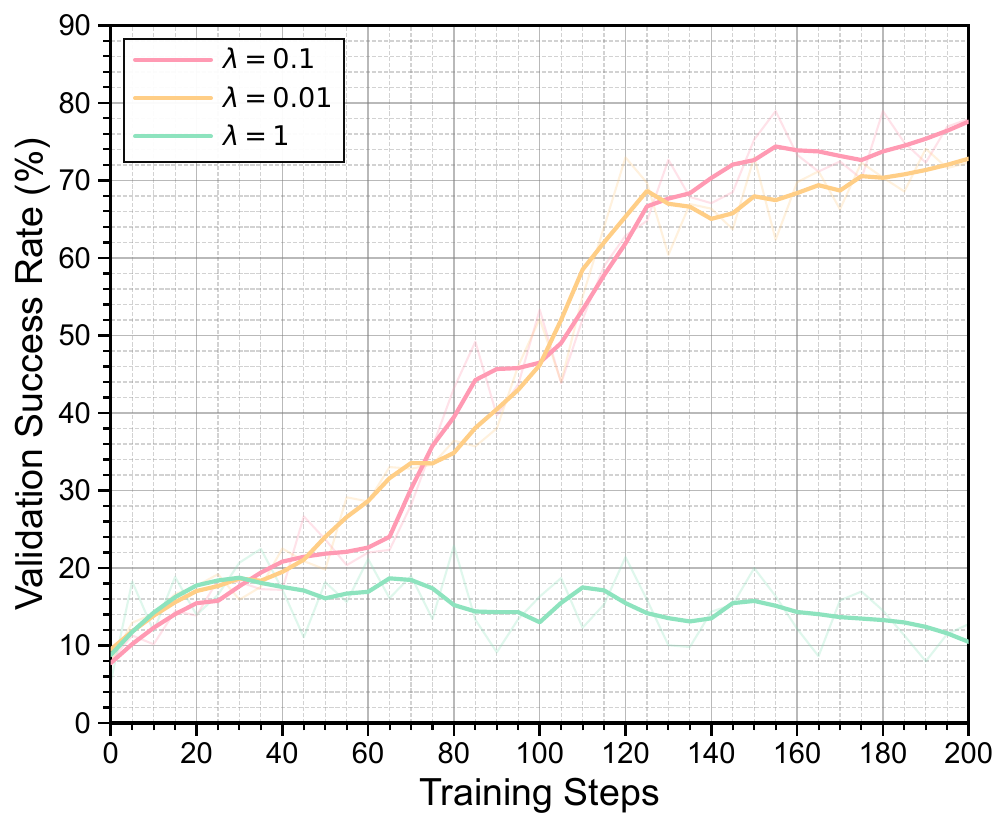}
\caption{Sensitivity of validation success rate to distillation weight $\lambda$.}
\label{fig:lambda}
\end{figure} 

\subsection{Sensitivity to Distillation Weight $\lambda$.}
Figure~\ref{fig:lambda} sweeps the distillation coefficient
$\lambda$ on Qwen2.5-1.5B-Instruct, revealing a clear trade-off
between the RL and distillation objectives.
At $\lambda = 1$, the distillation term dominates the gradient
and effectively freezes the policy near its initial performance
($\sim$15\%), preventing autonomous exploration beyond the
scaffold-prescribed behavior.
At $\lambda = 0.01$, the policy recovers RL-driven improvement
but internalizes scaffold-induced patterns too slowly, converging
roughly 5 points below the best setting.
The moderate value $\lambda = 0.1$ balances both pressures,
sustaining steady improvement to approximately 80\%---enough
distillation to accelerate internalization without suppressing
the RL objective's exploratory signal.
We adopt this value for all remaining experiments.

\subsection{Case Study: Experience Bank Evolution Dynamics}
\label{appendix:bank_evolution}
 
This section complements the aggregate gain-tracking curves in
Figure~\ref{fig:gain-tracking} with entry-level evidence from
ALFWorld training logs (Qwen2.5-7B-Instruct).
We trace four representative cases
(Table~\ref{tab:bank_case_summary}), each linking a rollout
failure to the experience it produces, its later retrieval, and
the resulting change in agent behavior.
Cases~1--2 (Figure~\ref{fig:bank_cases_early}) illustrate the
expansion phase, while Cases~3--4
(Figure~\ref{fig:bank_cases_late}) illustrate late-stage
refinement and non-stationary utility.

\begin{figure*}[t]
\begin{evolutioncase}{Case 1: From destination-first wandering to
object-first execution}
\textbf{Failure (step 150).}\;
The task is \texttt{put a clean bowl in shelf}.  The failed rollout
immediately navigates to the destination and loops around empty shelves:
 
{\small\ttfamily
go to shelf 3 $\rightarrow$ go to cabinet 1 $\rightarrow$
go to shelf 3 $\rightarrow$ go to cabinet 3 $\rightarrow$
examine shelf 3.}
 
\textbf{Successful contrast.}\;
The agent first searches likely sources, finds a bowl in the fridge, takes
and cleans it, then navigates to a shelf:
 
{\small\ttfamily
go to fridge 1 $\rightarrow$ open fridge 1 $\rightarrow$
take bowl 1 from fridge 1 $\rightarrow$
clean bowl 1 with sinkbasin 1 $\rightarrow$ go to shelf 1.}
 
\textbf{Reflected experience.}\;
Entry \texttt{task\_943}: \emph{``First identify where the target bowl is,
retrieve it, clean it if the goal requires a clean bowl, and only then move
it to a shelf.''}
 
\textbf{Later retrieval.}\;
At step 155, retrieved for \texttt{put a clean bowl in diningtable}
(similarity $0.886$).  The rollout no longer starts by visiting the table;
it locates, cleans, and places the bowl.  Utility: $U{=}0.125$.
\end{evolutioncase}
 
\vspace{4pt}
 
\begin{evolutioncase}{Case 2: From hallucinated source to source
verification}
\textbf{Failure (step 58).}\;
The task is \texttt{put some pillow on sofa}.  The agent goes to the sofa,
observes \texttt{box}, \texttt{creditcard}, \texttt{keychain}, and
\texttt{newspaper}---but no pillow.  It issues
\texttt{take pillow from sofa 1}, receives \texttt{Nothing happens}, and
repeats similar invalid source assumptions.
 
\textbf{Successful contrast.}\;
The paired trajectory searches alternative sources, reaches
\texttt{armchair 1}, observes \texttt{pillow 1}, takes it, and moves it to
the sofa---only issuing \texttt{take} when the observation lists the target.
 
\textbf{Reflected experience.}\;
Entry \texttt{step\_1067}: \emph{``Only use \texttt{take <object> from
<container>} when the object is actually listed at that location; if not
observed, do not repeat the same invalid action.''}
 
\textbf{Later retrieval.}\;
At step 60, retrieved for \texttt{put some keychain on ottoman}, where the
observation lacks the keychain---the same structural error.  Utility evolves
from $0.000$ at creation to $0.578$ at step 200 after 78 retrievals.
\end{evolutioncase}
\caption{\textbf{Experience bank evolution: early-stage expansion.}
Case~1 illustrates destination-first wandering corrected by an object-first
scaffold; Case~2 illustrates hallucinated source actions corrected by an
observation-gated rule.}
\label{fig:bank_cases_early}
\end{figure*}
 
\begin{figure*}[t]
\begin{evolutioncase}{Case 3: From unstructured Pick2 search to systematic
collection}
\textbf{Failure (step 165).}\;
For \texttt{find two ladle and put them in drawer}, the policy spends
actions on destination drawers or repeated navigation before confirming where
the two targets are.  Unlike early scaffolds, this failure involves managing
object count, source search, and final placement simultaneously.
 
\textbf{Reflected experience.}\;
Entry \texttt{task\_995}: \emph{``Identify likely locations of the target
item, inspect those locations methodically, pick up each required object,
and then place them into the specified container.  Avoid random movement or
opening unrelated containers before locating the targets.''}
 
\textbf{Later retrieval.}\;
At step 166, retrieved for \texttt{find two spatula and put them in drawer}
(similarity $0.854$).  The rollout locates \texttt{spatula 3} on a
countertop, picks it up, checks other locations, and deposits it in a
drawer.  Utility reaches $U{=}0.375$ by steps 195--200.
\end{evolutioncase}
 
\vspace{4pt}
 
\begin{evolutioncase}{Case 4: Utility tracking separates useful retrieval
from harmful repetition}
\textbf{Creation (step 144).}\;
Entry \texttt{step\_914}, \emph{``Avoid repeating a no-op movement,''} is
reflected from a \texttt{find two tissuebox and put them in drawer} failure.
The rule: if the agent is already at the target location, it should inspect
or act rather than re-issuing the same navigation action.
 
\textbf{Non-stationary utility.}\;
The entry is retrieved frequently but its utility is initially
unstable: $U{=}{-}0.026$ at step 150 and ${-}0.085$ at step 165---hovering
near the pruning threshold $\eta{=}{-}0.1$ but remaining above it.
Frequency-based retention would treat this entry as important despite its
near-zero or negative marginal gain; conversely, a positive threshold
would have discarded it prematurely.
 
\textbf{Recovery.}\;
As the policy reaches more states where repeated no-op movements are the
dominant error, the entry becomes useful: utility turns positive at
$U{=}0.048$ by step 180 and rises to $U{=}0.452$ at step 200 after 142
retrievals.  This illustrates why \methodname{} combines smoothed EMA
tracking (Eq.~\eqref{eq:ema}) with a mildly negative pruning threshold:
experience utility can shift as the policy distribution changes, and
premature pruning would discard entries whose value has not yet materialized.
\end{evolutioncase}
\caption{\textbf{Experience bank evolution: late-stage refinement and
non-stationary utility.}
Case~3 illustrates specialized multi-object coordination guidance;
Case~4 illustrates an entry whose utility is initially near the pruning
threshold but recovers as the policy's failure distribution shifts.}
\label{fig:bank_cases_late}
\end{figure*}
 
The four cases reveal a natural curriculum driven by the
policy's evolving failure distribution.
Early failures are structural and broadly shared: destination-first
wandering (Case~1) and hallucinated object presence (Case~2)
affect many task variants, producing generic scaffolds with
sustained high utility---these entries drive the initial bank
growth visible in Figure~\ref{fig:gain-tracking}.
As the policy internalizes these broad strategies, the remaining
failures narrow in scope.
By step~165, single-object manipulation is reliable but
coordinating two targets remains fragile;
Case~3's search-then-collect scaffold addresses precisely this
gap, and its moderate final utility ($U{=}0.375$) is consistent
with Pick2 remaining the hardest subtask in
Table~\ref{tab:ablation}.

Case~4 provides the most direct evidence for non-stationary
utility.
Entry \texttt{step\_914} (``avoid repeating no-op movements'')
hovers near the pruning boundary for roughly 20 steps
($U{=}{-}0.026$ at step~150, ${-}0.085$ at step~165) before
recovering to $U{=}0.452$ by step~200.
The mechanism is interpretable: once destination-first errors
(Case~1) are resolved, no-op navigation loops become the
dominant Pick2 failure mode, re-activating a previously marginal
entry.
A positive pruning threshold would have discarded it; the
adopted $\eta{=}{-}0.1$ retains such latent-value entries until
the policy's distribution shifts in their favor.

Together, these cases explain the aggregate bank trajectory in
Figure~\ref{fig:gain-tracking}---initial growth as generic
scaffolds accumulate, followed by contraction as the policy
absorbs broad patterns and the bank converges to a compact set
of specialized, currently relevant entries.

\section{Pseudocode}
Algorithm~\ref{alg:edge} outlines the full EDGE training loop,
which alternates among three stages per iteration:
experience-guided exploration scaffolding (\S\ref{sec:scaffold}),
gain-gated privileged distillation (\S\ref{sec:distill}),
and experience bank evolution (\S\ref{sec:bank}).

\begin{algorithm*}[t]
\renewcommand{\algorithmicrequire}{\textbf{Input:}}
\renewcommand{\algorithmicensure}{\textbf{Output:}}
\caption{EDGE: Experience-Distillation for Guided Exploration}
\label{alg:edge}
\begin{algorithmic}[1]
\Require
\Statex Policy $\pi_\theta$; experience bank $\mathcal{E}$;
\Statex Rollout group size $G$; distillation weight $\lambda$;
\Statex EMA momentum $\mu$; pruning threshold $\eta$;
\Statex Success-rate threshold $\xi$.

\For{each training iteration}
    \For{each task $x$ in batch}
        \State \textbf{\textit{Stage 1: Experience-Guided Exploration Scaffolding}}
        \State Retrieve top experience $e \in \mathcal{E}$ by embedding similarity.
        \State Sample $G$ trajectories from $\pi_\theta$:
        \Statex \quad\quad $\mathcal{T}^{\mathsf{T}}$ ($G/2$): conditioned on $c^{\mathsf{T}} = x \oplus e$
        \Statex \quad\quad $\mathcal{T}^{\mathsf{S}}$ ($G/2$): conditioned on $c^{\mathsf{S}} = x$
        \State Compute marginal gain $\Delta_e$ via Eq.~\eqref{eq:gain}.

        \State \textbf{\textit{Stage 2: Gain-Gated Privileged Distillation}}
        \If{$\Delta_e > 0$}
            \State Compute advantages over all $G$ trajectories.
            \For{each $\tau \in \mathcal{T}^{\mathsf{S}}$}
                \State Forward $\pi_{\mathrm{sg}(\theta)}$ on $\tau$ with $c^{\mathsf{T}}$.
                \State Compute $\mathcal{L}_{\text{distill}}$ via Eq.~\eqref{eq:distill}.
            \EndFor
        \Else
            \State Compute advantages over $\mathcal{T}^{\mathsf{S}}$ only.
            \State $\mathcal{L}_{\text{distill}} \leftarrow 0$.
        \EndIf
        \State $\mathcal{L}_{\text{actor}} \leftarrow \mathcal{L}_{\text{RL}} + \lambda\, \mathcal{L}_{\text{distill}}$.

        \State \textbf{\textit{Stage 3: Experience Bank Evolution}}
        \State Update EMA utility: $U_e^{(t)} \!\leftarrow\! (1\!-\!\mu)\, U_e^{(t-1)} + \mu\, \Delta_e$.
    \EndFor
    \State Update $\pi_\theta$ with $\mathcal{L}_{\text{actor}}$.
    \State Prune experiences with $U_e^{(t)} < \eta$ from $\mathcal{E}$.
    \State Identify task categories with success rate $< \xi$.
    \State Generate new experiences via $f_{\text{reflect}}(\tau^{+}, \tau^{-})$; insert into $\mathcal{E}$.
\EndFor
\Ensure Trained policy $\pi_\theta$ (deployed without $\mathcal{E}$).
\end{algorithmic}
\end{algorithm*}

\section{Prompts}
This section provides the prompt templates used in our experiments. We include
both rollout prompts and experience update prompts for ALFWorld and WebShop.
The rollout prompts are used for action generation, while the update prompts
are used to generate new state-aware experiences from contrasted trajectories.

\subsection{ALFWorld}

Figure~\ref{fig:alfworld-rollout-prompts} shows the ALFWorld rollout prompts.
The first prompt provides retrieved experiences as additional context, while
the second prompt removes external experiences and asks the agent to act only
based on the current task, history, observation, and admissible actions.

\begin{figure*}[t]
\small
\begin{simplepromptbox}{Prompt: ALFWorld Agent Execution with Experience}

\textbf{System Prompt:}

You are an expert agent operating in the ALFRED Embodied Environment. Your task is to:
\placeholder{task\_description}

\vspace{3pt}
\textbf{\# Retrieved Relevant Experience}

\placeholder{retrieved\_experiences}

\textit{Warning: These experiences may be outdated. Use them only if they align with your current observation.}

\vspace{3pt}
\textbf{\# Current Progress}

Prior to this step, you have already taken \placeholder{step\_count} step(s).
Below are the most recent \placeholder{history\_length} observations and the
corresponding actions you took:
\placeholder{action\_history}

You are now at step \placeholder{current\_step} and your current observation is:
\placeholder{current\_observation}

Your admissible actions of the current situation are:
[\placeholder{admissible\_actions}].

\vspace{3pt}
Now it is your turn to take an action. You should first reason step-by-step about
the current situation. This reasoning process \textbf{MUST} be enclosed within
\texttt{<think>} \texttt{</think>} tags. Once you have finished your reasoning,
you should choose an admissible action for the current step and present it within
\texttt{<action>} \texttt{</action>} tags.

\end{simplepromptbox}

\vspace{5pt}

\begin{simplepromptbox}{Prompt: ALFWorld Agent Execution without Experience}

\textbf{System Prompt:}

You are an expert agent operating in the ALFRED Embodied Environment. Your task is to:
\placeholder{task\_description}

\vspace{3pt}
\textbf{\# Retrieved Relevant Experience}

No external general and task-specific experiences are provided. Use your learned
strategy and current observation.

\vspace{3pt}
\textbf{\# Current Progress}

Prior to this step, you have already taken \placeholder{step\_count} step(s).
Below are the most recent \placeholder{history\_length} observations and the
corresponding actions you took:
\placeholder{action\_history}

You are now at step \placeholder{current\_step} and your current observation is:
\placeholder{current\_observation}

Your admissible actions of the current situation are:
[\placeholder{admissible\_actions}].

\vspace{3pt}
Now it is your turn to take an action. You should first reason step-by-step about
the current situation. This reasoning process \textbf{MUST} be enclosed within
\texttt{<think>} \texttt{</think>} tags. Once you have finished your reasoning,
you should choose an admissible action for the current step and present it within
\texttt{<action>} \texttt{</action>} tags.

\end{simplepromptbox}

\caption{ALFWorld rollout prompts with and without retrieved experience.}
\label{fig:alfworld-rollout-prompts}
\end{figure*}

Figure~\ref{fig:experience-update-prompt} shows the prompt used to update the
ALFWorld experience bank. Given a failed trajectory and a successful reference
trajectory for the same task, the model generates compact state-aware
experiences in JSON format.
\begin{figure*}[t]
\small
\begin{simplepromptbox}{Prompt: ALFWorld Experience Update}

\textbf{System Prompt:}

You are an expert updating an ALFWorld state-aware experience bank. You are
given one failed trajectory and one successful trajectory for the same task.
Analyze the contrasted rollout snippets below and propose \textbf{NEW or revised
experiences} that help the agent act better in the same environment state.

\vspace{3pt}
\textbf{\# Task Information}

Task: \placeholder{task\_text}

Task Type: \placeholder{task\_type}

The task was \placeholder{successfully/unsuccessfully} completed.

\vspace{3pt}
\textbf{\# Contrasted Rollout Snippets}

Failed trajectory:
\placeholder{failed\_text}

Successful trajectory (reference):
\placeholder{success\_text}

\vspace{3pt}
\textbf{\# Requirements}

\begin{itemize}[leftmargin=1.4em, itemsep=1pt, topsep=1pt, parsep=0pt]
    \item Each experience must be \textbf{state-aware}, not a generic tip.
    \item Focus on what the successful no-experience rollout did, and what the
    retrieved experience may have caused the agent to do incorrectly.
    \item Prefer compact trigger language that can be matched at retrieval time.
    \item Use actionable wording tied to admissible actions and visible state cues.
    \item Return only JSON.
\end{itemize}

Generate 1--\placeholder{max\_new\_skills\_per\_update} new experiences.

\vspace{3pt}
\textbf{\# Output Format Example}

\begin{promptjson}
{
  "title": "Plan object location before acting",
  "principle": "For this task type, first identify where the object is, then plan the sequence of actions.",
  "when_to_apply": "When the task involves finding or moving a specific object"
}
\end{promptjson}

\end{simplepromptbox}

\caption{Prompt for updating the ALFWorld state-aware experience bank from
contrasted failed and successful trajectories. The model is instructed to
generate compact, state-aware experiences that can guide future retrieval and
decision making.}
\label{fig:experience-update-prompt}
\end{figure*}

\subsection{WebShop}

Figure~\ref{fig:webshop-rollout-prompts} shows the WebShop rollout prompts.
The experience-conditioned prompt includes retrieved memories, while the
experience-free prompt relies only on the shopping instruction, interaction
history, current observation, and admissible actions.

\begin{figure*}[t]
\small
\begin{simplepromptbox}{Prompt: WebShop Agent Execution with Experience}

\textbf{System Prompt:}

You are an expert autonomous agent operating in the WebShop e-commerce environment.
Your task is to: \placeholder{task\_description}.

\vspace{3pt}
\textbf{\# Retrieved Relevant Experience}

\placeholder{retrieved\_memories}

\textit{Warning: These experiences may be outdated. Use them only if they align with your current observation.}

\vspace{3pt}
\textbf{\# Current Progress}

Prior to this step, you have already taken \placeholder{step\_count} step(s).
Below are the most recent \placeholder{history\_length} observations and the
corresponding actions you took: \placeholder{action\_history}

You are now at step \placeholder{current\_step} and your current observation is:
\placeholder{current\_observation}.

Your admissible actions of the current situation are:
\placeholder{available\_actions}

Now it is your turn to take one action for the current step.
You should first reason step-by-step about the current situation, then think
carefully which admissible action best advances the shopping goal. This
reasoning process \textbf{MUST} be enclosed within
\texttt{<think>} \texttt{</think>} tags.
Once you have finished your reasoning, you should choose an admissible action
for current step and present it within
\texttt{<action>} \texttt{</action>} tags.

\end{simplepromptbox}

\vspace{5pt}

\begin{simplepromptbox}{Prompt: WebShop Agent Execution without Experience}

\textbf{System Prompt:}

You are an expert autonomous agent operating in the WebShop e-commerce environment.
Your task is to: \placeholder{task\_description}.

\vspace{3pt}
\textbf{\# Retrieved Relevant Experience}

No external general and task-specific experiences are provided. Use your learned
strategy and current observation.

\vspace{3pt}
\textbf{\# Current Progress}

Prior to this step, you have already taken \placeholder{step\_count} step(s).
Below are the most recent \placeholder{history\_length} observations and the
corresponding actions you took: \placeholder{action\_history}

You are now at step \placeholder{current\_step} and your current observation is:
\placeholder{current\_observation}.

Your admissible actions of the current situation are:
\placeholder{available\_actions}

Now it is your turn to take one action for the current step.
You should first reason step-by-step about the current situation, then think
carefully which admissible action best advances the shopping goal. This
reasoning process \textbf{MUST} be enclosed within
\texttt{<think>} \texttt{</think>} tags.
Once you have finished your reasoning, you should choose an admissible action
for current step and present it within
\texttt{<action>} \texttt{</action>} tags.

\end{simplepromptbox}

\caption{WebShop rollout prompts with and without retrieved experience.}
\label{fig:webshop-rollout-prompts}
\end{figure*}

Figure~\ref{fig:webshop-experience-update-prompt} shows the prompt used to
update the WebShop experience bank. The model compares failed and successful
shopping trajectories and produces new state-aware experiences tied to visible
state cues and available actions.

\begin{figure*}[t]
\small
\begin{simplepromptbox}{Prompt: WebShop Experience Update}

\textbf{System Prompt:}

You are an expert updating a WebShop shopping state-aware experience bank. You
are given one failed trajectory and one successful trajectory for the same task.
Analyze the contrasted rollout snippets below and propose \textbf{NEW or revised
experiences} that help the agent act better in the same environment state.

\vspace{3pt}
\textbf{\# Task Information}

Task: \placeholder{task\_text}

Task Type: \placeholder{task\_type}

The task was \placeholder{successfully/unsuccessfully} completed.

\vspace{3pt}
\textbf{\# Contrasted Rollout Snippets}

Failed trajectory:
\placeholder{failed\_text}

Successful trajectory (reference):
\placeholder{success\_text}

\vspace{3pt}
\textbf{\# Requirements}

\begin{itemize}[leftmargin=1.4em, itemsep=1pt, topsep=1pt, parsep=0pt]
    \item Each experience must be \textbf{state-aware}, not a generic tip.
    \item Focus on what the successful no-experience rollout did, and what the
    retrieved experience may have caused the agent to do incorrectly.
    \item Prefer compact trigger language that can be matched at retrieval time.
    \item Use actionable wording tied to available actions and visible state cues.
    \item Return only JSON.
\end{itemize}

Generate 1--\placeholder{max\_new\_skills\_per\_update} new experiences.

\vspace{3pt}
\textbf{\# Output Format Example}

\begin{promptjson}
{
  "title": "Verify Early, Abort Fast",
  "principle": "On the product page, immediately check category, core attributes, and price; if a key constraint is violated, leave the page at once.",
  "when_to_apply": "Within the first observation on every product detail page."
}
\end{promptjson}

\end{simplepromptbox}

\caption{Prompt for updating the WebShop shopping state-aware experience bank
from contrasted failed and successful trajectories. The model is instructed to
generate compact, state-aware experiences that can guide future retrieval and
shopping decisions.}
\label{fig:webshop-experience-update-prompt}
\end{figure*}

\section{Dataset License}
Our experiments are based on the publicly available ALFWorld~\citep{DBLP:conf/iclr/ShridharYCBTH21} and WebShop~\citep{DBLP:conf/nips/Yao0YN22} environments. 
Training, evaluation, and experience-bank construction are performed using trajectories generated from the task instances and interaction interfaces provided by these environments. 
We strictly follow the licenses and usage terms of ALFWorld and WebShop, and use the resulting data only for academic research purposes. 
No private, personally identifiable, or proprietary user data is used in any stage of training, evaluation, or experience construction.

\section{LLMs Usage Statement}
We employed a Large Language Model (LLM) to assist exclusively in the editorial stage of manuscript preparation. Its role was limited to refining phrasing, correcting grammar, and enhancing clarity and readability across different sections. The LLM had no involvement in formulating research ideas, designing experiments, or conducting analyses. All scientific contributions and findings are entirely the work of the authors. The authors have ensured that the use of the LLM complies with ethical standards, avoiding plagiarism and scientific misconduct.

\end{document}